\documentclass[10pt,twocolumn,letterpaper]{article}

\usepackage{cvpr}
\usepackage{times}
\usepackage{epsfig}
\usepackage{graphicx}
\usepackage{amsmath}
\usepackage{amssymb}
\usepackage{enumitem}

\usepackage{booktabs}
\usepackage{multirow}
\usepackage{graphicx}
\usepackage{tablefootnote}

\usepackage[pagebackref=true,breaklinks=true,letterpaper=true,colorlinks,bookmarks=false]{hyperref}

\cvprfinalcopy 

\def\cvprPaperID{5483} 

\ifcvprfinal

\begin{document}

\title{Multi-Modal Graph Neural Network for Joint Reasoning \\ on Vision and Scene Text}

\author{Difei Gao\textsuperscript{1,2*}, Ke Li\textsuperscript{1,2*}, Ruiping Wang\textsuperscript{1,2}, Shiguang Shan\textsuperscript{1,2}, Xilin Chen\textsuperscript{1,2}\\
\textsuperscript{1}Key Laboratory of Intelligent Information Processing of Chinese Academy of Sciences (CAS),\\
Institute of Computing Technology, CAS, Beijing, 100190, China\\
\textsuperscript{2}University of Chinese Academy of Sciences, Beijing, 100049, China\\
{\tt\small \{difei.gao, ke.li\}@vipl.ict.ac.cn, \{wangruiping, sgshan, xlchen\}@ict.ac.cn
}}

\maketitle
\thispagestyle{empty}

\begin{abstract}
Answering questions that require reading texts in an image is challenging for current models. One key difficulty of this task is that rare, polysemous, and ambiguous words frequently appear in images, \eg names of places, products, and sports teams. To overcome this difficulty, only resorting to pre-trained word embedding models is far from enough. A desired model should utilize the rich information in multiple modalities of the image to help understand the meaning of scene texts, \eg the prominent text on a bottle is most likely to be the brand. 
Following this idea, we propose a novel VQA approach, Multi-Modal Graph Neural Network (\textbf{MM-GNN}). It first represents an image as a graph consisting of three sub-graphs, depicting visual, semantic, and numeric modalities respectively. Then, we introduce three aggregators which guide the message passing from one graph to another to utilize the contexts in various modalities, so as to refine the features of nodes. The updated nodes have better features for the downstream question answering module. Experimental evaluations show that our \textbf{MM-GNN} represents the scene texts better and obviously facilitates the performances on two VQA tasks that require reading scene texts.

\end{abstract}
{\let\thefootnote\relax\footnotetext{* indicates equal contribution.}} \par


\section{Introduction}

The texts in a scene convey rich information that is crucial for performing daily tasks like finding a place, acquiring information about a product, etc. An advanced Visual Question Answering (VQA) model which is able to reason over scene texts and other visual contents could have extensive applications in practice, such as assisting visually impaired users, and education of children. Our focus in this paper is to endow VQA models the ability of better representing the image containing the scene texts, to facilitate the performances of answering on VQA tasks~\cite{singh2019towards, biten2019scene} that requires reading in images.

\begin{figure}
\centering
\includegraphics[height=6.5cm]{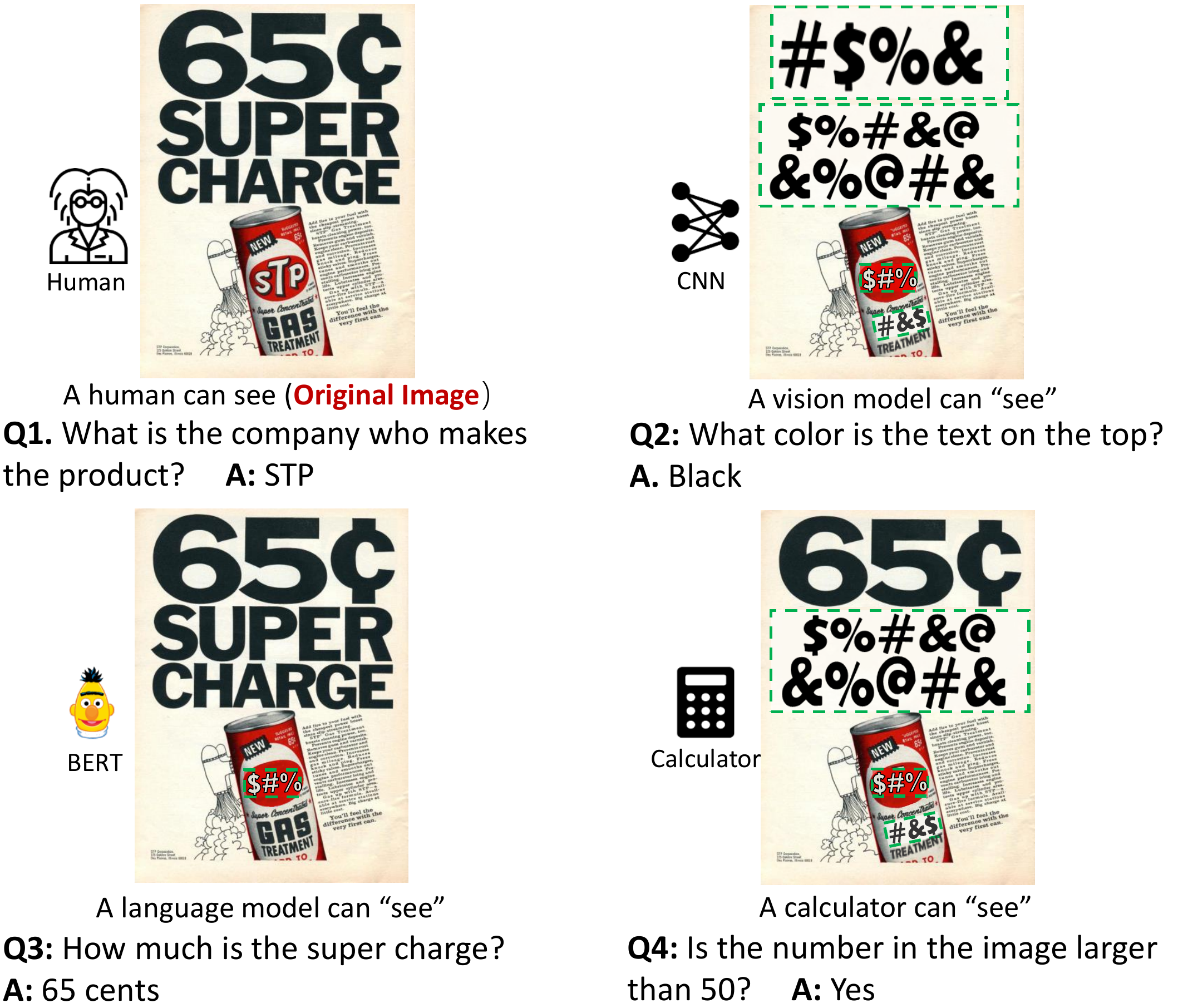}
\caption{
An image could contain information in multiple modalities, thus it looks different to models with different abilities. For example, the image in the eye of a human (top left) combines multi-modal contents. The visual modality contains the visual appearances of objects and texts. The semantic modality involves the semantics of the texts, yet it cannot determine the semantics of rare words like ``STP'' in the image. The numeric modality is about the numeric relation between numbers, like 65 is larger than 50. \textbf{Q2} to \textbf{Q4} are three common questions involving reasoning on one of these modalities; while \textbf{Q1} requires using visual context to infer the semantic of ``STP''. Random characters within green dashed boxes represent modalities out of the observer's capability.}
\label{fig1}
\vspace{-0.5cm}
\end{figure}

\begin{figure}
\centering
\includegraphics[height=3.0cm]{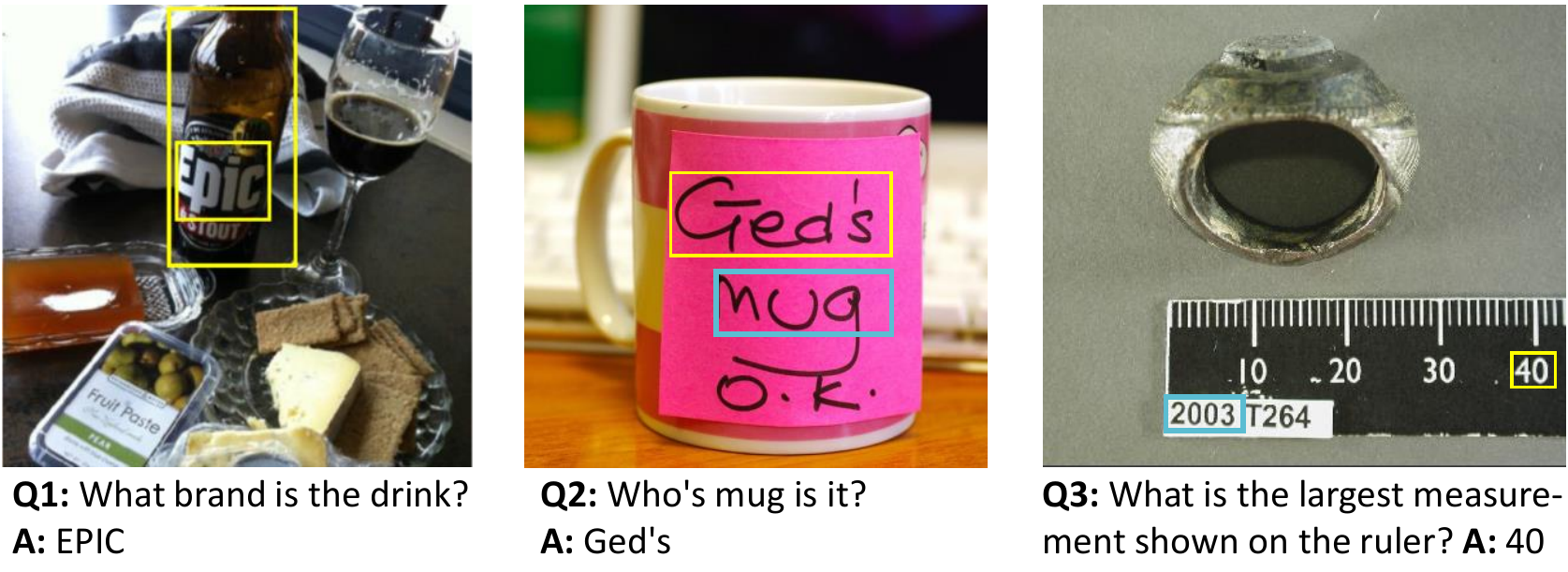}
\caption{Three example questions requiring leveraging different types of multi-modal contexts to answer the questions. \textbf{Q1:} the model should use visual context (the ``EPIC'' is a prominent word on the bottle) to infer the semantics of the word ``EPIC''. \textbf{Q2:} the model should infer that ``Ged'' indicates the owner of the mug by using the semantic context of the word. \textbf{Q3:} The model should be able to utilize the semantic of the numbers to depict more informative numeric relations between numbers, such as ``40'' is the largest measurement among ``40'', ``30''.}

\label{fig1.2}
\vspace{-0.5cm}
\end{figure}

What are the unique challenges of modeling scene texts compared to the pure visual entities (such as objects and scenes) and the natural language texts (sentences or phrases)? A scene text inherently contains information in multiple modalities, {\em visual information}, including color, shape, and {\em semantic information}, \eg ``New York'' is the name of a city, and {\em numeric information} for numbers, \eg ``65'' is larger than ``50''. These types of information are frequently used in answering daily questions. For example in Fig.~\ref{fig1}, \textbf{Q2} requires the model to find the target scene text with its visual information; \textbf{Q3} needs the model to understand the semantic of ``65'' which indicates the amount of the money; \textbf{Q4} requires the understanding of numeric relation between numbers. Therefore, to correctly answer the questions involving scene texts, it is indispensable to clearly depict each modality of the scene texts. In addition, among these three modalities, it is more difficult to determine the semantics of scene texts, because the scene texts encountered in daily environments have a large possibility to be unknown, rare or polysemous words, \eg, the name of a product ``STP'' as shown in Fig.~\ref{fig1}. To tackle this problem, the model should be able to determine the semantics of these texts beyond only using word embedding~\cite{pennington2014glove, joulin2017bag} pre-trained on a text corpus. In this paper, we propose to teach the model how to utilize the context in different modalities in surrounding of the words to determine their meanings like a human, that is, 1) visual context: the prominent word on the bottle is most likely to be its brand, as \textbf{Q1} in Fig.~\ref{fig1} and \textbf{Q1} in Fig.~\ref{fig1.2}, 2) semantic context: the surrounding texts of a rare or ambiguous word may help to infer its meaning, \eg \textbf{Q2} shown in Fig.~\ref{fig1.2}. In addition, utilizing semantics of numbers can also depict more informative numeric relations between numbers, as \textbf{Q3} shown in Fig.~\ref{fig1.2}.

Following the aforementioned ideas, we propose a novel approach, Multi-Modal Graph Neural Networks (MM-GNN), to obtain a better representation of the multi-modal contents in an image and facilitate question answering. Our proposed MM-GNN contains three sub-graphs for representing three modalities in an image, i.e., visual modality for visual entities (including texts and objects), semantic modality for scene texts, and numeric modality for number-related texts, as shown in Fig.~\ref{model}.
The initial representations of nodes in three graphs are obtained from priors, such as word embedding learned from the corpora and Faster R-CNN features.
Then, MM-GNN dynamically updates the representations of nodes by three attention-based aggregators, corresponding to utilizing three typical types of contexts in Fig.~\ref{fig1.2}. These aggregators calculate the relevance scores of two nodes considering their visual appearances and layout information in the image, together with questions. Besides relevance between nodes, by attending on the basis of layout information, we are actually linking texts to their \emph{physical carriers} (the object a text is printed or carved on); and given language hints, attention models can pass messages more accurately, by considering the directives implied by questions. Three different aggregators guide the message passing from one modality to another modality (or to itself) to leverage different types of contexts to refine the node features in a certain order. The updated representation contains richer and more precise information, facilitating the answering model to attend to the correct answer.

Finally, we conduct experiments with our proposed MM-GNN and its variants on two recently proposed datasets TextVQA~\cite{singh2019towards} and ST-VQA~\cite{biten2019scene}. The results show that our MM-GNN with newly designed aggregators effectively learns the representations of the scene texts and facilitates the performance of VQA tasks that require reading texts. 

\section{Related Work}

\textbf{Visual Question Answering Tasks.}
In recent years, numerous works have proposed diverse VQA tasks~\cite{ren2015exploring, malinowski2014multi, antol2015vqa, goyal2016making, wang2018fvqa, shah2019kvqa, zellers2019recognition, johnson2017clevr, hudson2019gqa} for evaluating different types of core skills for answering visual questions. One line of datasets~\cite{ren2015exploring, malinowski2014multi, antol2015vqa, goyal2016making}, such as COCO-QA and VQA, studies questions about querying the visual information of an image. Relevant works~\cite{lu2016hierarchical, fukui2016multimodal, schwartz2017high, anderson2017bottom, ben2017mutan, noh2016image, yang2016stacked} propose various attention mechanisms and multi-modal fusion techniques to better locate the image region for a given question to facilitate the answering procedure. Another line of works, such as CLEVR and GQA, introduces questions demanding complex and compositional spatial reasoning skills. Relevant works on these tasks introduce modular networks~\cite{andreas2016learning, andreas2016neural, hu2017learning, johnson2017inferring, hudson2018compositional} and neural-symbolic model~\cite{shi2019explainable, yi2018neural} which can robustly generate answer by performing explicit multi-step reasoning on an image.

In this paper, we focus on a new type of questions recently proposed by the TextVQA~\cite{singh2019towards} and ST-VQA~\cite{biten2019scene}. Compared to other VQA tasks, these two tasks are unique in introducing questions about images that contain multi-modal contents, including visual objects and diverse scene texts. 
To solve these tasks, this paper focuses on how to formulate the multi-modal contents and obtain better representations of scene texts and objects.


\textbf{Representation Learning in VQA.}
Some inspiring works have studied the representation of images to improve the performance of VQA tasks. The VQA models~\cite{lu2016hierarchical, fukui2016multimodal, schwartz2017high, noh2016image, yang2016stacked} in the early stage mainly use the VGG or ResNet feature pre-trained on the ImageNet to represent images. However, this type of grid-level feature is limited to perform object-level attention. Therefore, ~\cite{anderson2017bottom} proposes to represent one image as a list of detected object features. Besides, to solve complex compositional questions, ~\cite{shi2019explainable, yi2018neural} propose some symbolic structural representations of the synthetic images (\eg a scene graph extracted from an image) in CLEVR which allow a VQA system to perform explicit symbolic reasoning on them. More recently, ~\cite{norcliffe2018learning, li2019relation, hu2019language} represent the natural image as a fully connected graph (can be viewed as an implicit scene graph where the relations between objects are not explicitly represented). This type of graph allows the model to predict dynamic edge weights to focus on a sub-graph related to the question and is widely used in natural images QA.

The above-mentioned methods all focus on the representation of visual objects, while this paper extends it to represent images with multi-modal contents. We represent one image as a graph composed of three sub-graphs to separately depict the entities in each modality and build the connections between entities in different modalities.

\textbf{Graph Neural Network.} 
Graph Neural Network (GNN)~\cite{scarselli2008graph, bruna2014spectral, kipf2016semi, velivckovic2017graph, xu2018powerful} is a powerful framework for representing graph-structured data. 
The GNN follows an aggregation scheme that controls how the representation vector of a node calculated by its neighboring nodes to capture specific patterns of a graph. 
Recently, numerous variants of GNN are proposed to capture different types of patterns of the graph in many tasks. For graph classification tasks, many works on text classification~\cite{scarselli2008graph, velivckovic2017graph, defferrard2016convolutional}, and protein interface prediction~\cite{fout2017protein} utilize the GNN to iteratively combine the information of the neighboring nodes to capture the structure information of the graph.

In addition, many interesting works~\cite{teney2017graph, norcliffe2018learning, li2019relation, hu2019language, wang2019neighbourhood} introduce GNN for grounding related task, such as referring expression~\cite{kazemzadeh2014referitgame} and visual question answering~\cite{zhang2016yin, goyal2016making, hudson2019gqa}. These works~\cite{teney2017graph, norcliffe2018learning, li2019relation, hu2019language, wang2019neighbourhood} propose GNN with language conditioned aggregator to dynamically locate a sub-graph of the scene for a given query (e.g. a referring expression or a question), then GNN updates the features of the nodes in the sub-graph to encode the relations among objects. The updated nodes have better features for latter grounding related tasks. 

Similar to the previous GNNs~\cite{teney2017graph, norcliffe2018learning, li2019relation, hu2019language, wang2019neighbourhood} for grounding related tasks, we utilize the GNN to obtain better features. But this paper extends GNN from performing reasoning on a single-modal graph to a multi-modal graph. Besides, our proposed new aggregation schemes can explicitly capture different types of multi-modal contexts to update the representation of the nodes.

\section{Method}
In this section, we elaborate on the proposed multi-modal graph neural networks (MM-GNN) for answering visual questions that require reading. Given an image, which contains visual objects and scene texts, and a question, the goal is to generate the answer. Our model answers the question in three steps: (1) extract the multi-modal contents of an image and construct a three-layer graph, (2) perform multi-step message passing among different modalities to refine the representation of the nodes, and (3) predict the answer based on the graph representation of the image.

\begin{figure*}
\centering
\includegraphics[height=3.9cm]{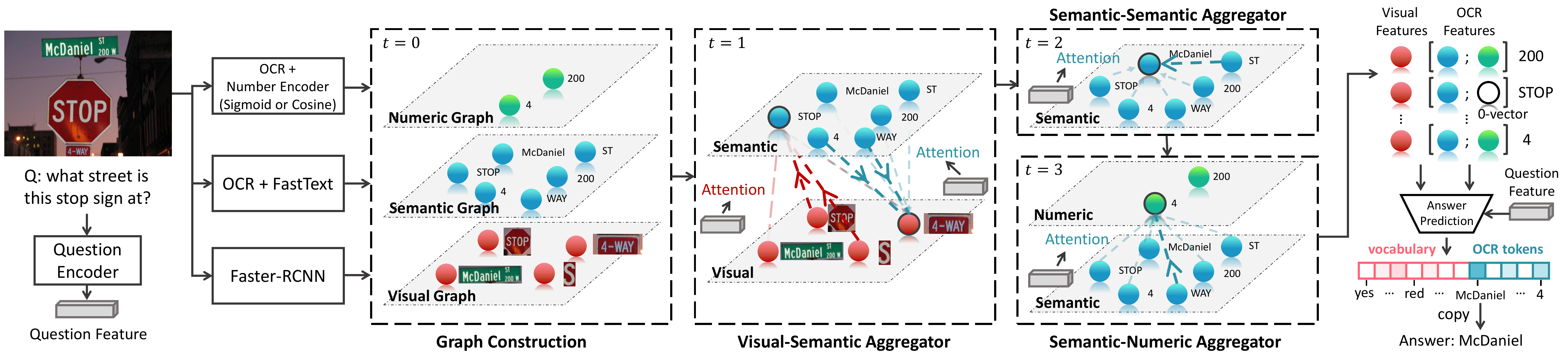}
\caption{The architecture of Multi-Modal Graph Neural Network (\textbf{MM-GNN}). It first runs three offline models to initialize the representation of each graph. After that, three aggregators successively update the representations of nodes by passing the information in an inter-graph or intra-graph way to obtain better representations of the nodes. Finally, the answer prediction module uses these features to output the answer. The blue or red arrows on lines between two nodes indicate the directions of information aggregation, with deeper lines representing higher attention. [;] indicates the concatenating operation.}
\label{model}
\end{figure*}

\subsection{Multi-Modal Graph Construction}
As shown in Fig.~\ref{mm-gnn}, given an image, we first construct a multi-modal graph composed of three sub-graphs, i.e., visual graph, semantic graph and numeric graph for representing the information in three modalities. The \emph{visual graph} $G_v$ is a fully connected graph, where each node $v_i \in {V_v} = \{v_i\}_{i=1}^N$ encodes the pure visual information of entities (i.e., objects and scene texts) and $N$ is the number of candidate objects generated by the extractor. The initial representation $v_i^{(0)}$ of $v_i$ is obtained by using an image feature extractor, \eg Faster R-CNN~\cite{girshick2015fast} detector. 

The \emph{semantic graph} $G_s$ is also a fully connected graph, and each node $s_i \in V_s = \{s_i\}_{i=1}^M$ represents the semantic meaning of a scene text, \eg ``New York'' is the name of a city, ``Sunday'' is one day in a week, and $M$ is the number of extracted tokens. Concretely, to obtain the semantic graph, we first use an Optical Character Recognition (OCR) model to extract word tokens in images. Then, the $i$-th token is embedded by a pre-trained word embedding model as the initial representation $s_i^{(0)}$ of node $s_i$.

Besides, for number-type strings, \eg ``2000'', they not only contain semantic meanings indicating string type, \eg year (or dollars), but also numeric meanings which indicate the numeric relations among other number-type strings, \eg ``2000'' is larger than ``1900''. 
Thus, we construct a fully connected \emph{numeric graph} $G_n$ to represent such information of number-type texts $x_i \in V_n = \{x_i\}_{i=1}^K$.  We categorize common numeric texts into several types, \eg number, time, etc. Then number-type texts are embedded into -1 to 1, denoted as $x_i^{(0)}$, with sigmoid function (for monotone number, like ``12'') or cosine function (for period number, like ``10:00'') according to their categories, where $K$ is the number of number-type texts. More details of the number encoder are in the Supp. Besides, the entire graph composed of three sub-graphs is overall fully connected, but only a specific part of nodes and edges is used in one aggregator.
 

\subsection{Aggregation Scheme}
After constructing the graph and initializing the representation of each node, we propose three aggregators which guide the information flow between one sub-graph to another or itself to utilize the different types of context to refine the representation of the nodes, as shown in Fig.~\ref{mm-gnn}.

\textbf{Visual-Semantic Aggregator.} The first aggregator is the Visual-Semantic aggregator, goal of which is two-fold: 1) leverage the visual context to refine a semantic node (for solving questions like \textbf{Q1} in Fig.~\ref{fig1.2}) and 2) utilize the semantic context to refine a visual node, making the visual representation of texts' physical carriers aware of the text (for solving questions like \textbf{Q3} in Fig.~\ref{mmgnn_vs_no_gnn}). Here, we first illustrate the implementation of the first goal. For each node $s_i$ in semantic graph $G_s$, the aggregator updates the representation of $s_i$ by first attending on relevant neighbour nodes in visual graph $v_j \in \mathcal{N}_{s_i}^v = \{v_j\}_{j=1}^N$, then aggregating the information of attended nodes to update the representation of $s_i$. Concretely, we first calculate the relevance score $a_{v_j,s_i}^s$ between the node $s_i$ and its neighboring node $v_j$ based on their visual representation and their location features $b_{s_i}$ and $b_{v_j}$ (i.e. the coordinates of bounding boxes)  and the question feature $q$ obtained by embedding the question words and going through an LSTM~\cite{hochreiter1997long}, formulated as,

\begin{align}\nonumber
a'_{v_j,s_i} &= f_s([s_i^{(0)}; f_b(b_{s_i})])^T (f_v([v_j^{(0)}; f_b(b_{v_{j}})]) \odot f_q(q)), \\
a_{v_j, s_i}^{s} &= \frac{\exp(a'_{v_j,s_i})}{\sum_{v_j \in \mathcal{N}_{s_i}^v} \exp(a'_{v_j,s_i})},
\end{align}
where $f_s$, $f_v$, $f_b$ and $f_q$ are the MLPs for encoding the semantic nodes, visual nodes, bounding boxes features, and question feature respectively, $[;]$ indicates concatenating two vectors, and $\odot$ is element-wise multiplication. Here, we also consider the question information in calculating the attention score, because we hope the model can aggregate the related nodes considering the information in the question.    Then, we aggregate the information of attended nodes, and append the aggregated features to $s_i^{(0)}$ depicting the additional information of this node to obtain the updated semantic representation, formulated as,
\begin{align}
s_i^{(1)} = [s_i^{(0)}; \sum_{v_j \in \mathcal{N}_{s_i}^v} a_{v_j,s_i}^s f_{v'}(v_j^{(0)})],
\end{align}
where $s_i^{(1)}$ is the updated node representation at $t$=$1$ (shown in Fig.~\ref{model}), and $f_{v'}$ is an MLP to encode the features of neighboring nodes. 

Similar to the scheme of refining semantic nodes, we obtain the updated representation of nodes $v^{(1)}_j$ in $G_v$ by
\begin{align}
a_{v_j, s_i}^v &= \frac{\exp(a'_{v_j,s_i})}{\sum_{s_i \in \mathcal{N}_{v_j}^s} \exp(a'_{v_j,s_i})}  \\
v_j^{(1)} &= [v_i^{(0)}; \sum_{s_i \in \mathcal{N}_{v_j}^s} a_{v_j,s_i}^v f_{s'}(s_j^{(0)})],
\end{align}
where $f_{s'}$ is an MLP to encode the $s_j$, and $\mathcal{N}_{v_j}^s$ indicates the neighboring nodes of $v_j$ in semantic graph. Note that in all aggregators, the additional information is appended after original features; specifically, after Visual-Semantic aggregation, the dimensions of both semantic and visual features are multiplied by two.

\textbf{Semantic-Semantic Aggregator.} This aggregator then refines the representation of each semantic node by considering its semantic context (for solving questions like \textbf{Q2} in Fig.~\ref{fig1.2}). For each node $s_i$, the aggregator finds the proper neighboring nodes in semantic graph $\mathcal{N}_{s_i}^s = \{s_j |$ $j \in \{1,..., M\}$ and $j \notin {i}\}$ by attention mechanism, then aggregating the information of attended nodes. 

More specifically, the relevance score $a_{s_j,s_i}$ of the node $s_i$ and its neighboring node $s_j$ is computed by their semantic representation and their location features $b_{s_i}$ and $b_{s_j}$ in images, formulated as,

\begin{align}\nonumber
a'_{s_j,s_i} &= g_{s_1}([s_i^{(1)}; g_b(b_{s_i})])^T (g_{s_2}([s_j^{(1)}; g_b(b_{s_j})]) \odot g_q(q)), \\
a_{s_j, s_i} &= \frac{\exp(a'_{s_j,s_i})}{\sum_{j \in \mathcal{N}_{s_i}^s} \exp(a'_{s_j,s_i})},
\end{align}
where $g_{s_1}$, $g_{s_2}$, $g_b$, and $g_q$ are the MLPs for encoding the node features (the first two), bounding boxes features and question features.
Then, we aggregate the information of attended nodes, and append the aggregated features to $s_i$ as,

\begin{align}
s_{i}^{(2)} = [s_i^{(1)}; \sum_{s_j \in \mathcal{N}_{s_i}^s} a_{s_j,s_i} g_{s_3}(s_j^{(1)})],
\end{align}
where $s_i^{(2)}$ is the updated node representation at $t$=$2$, and $g_{s_3}$ is an MLP to encode the features of neighboring nodes.

\textbf{Semantic-Numeric Aggregator.} The goal of this aggregator is to leverage the semantic context to refine the value nodes to depict more informative numeric relations between numbers (for solving questions like \textbf{Q3} in Fig.~\ref{fig1.2}). The mechanism of semantic-numeric aggregator is similar to the mechanism of achieving the first goal in Visual-Semantic aggregator. We first calculate the relevance score $a_{s_j, x_i}$ between nodes $s_j$ and $x_i$, then aggregate the information of semantic nodes to numeric nodes, formulated as,
\begin{align}
x_{i}^{(3)} = [x_i^{(0)}; \sum_{s_j \in \mathcal{N}_{x_i}^s} a_{s_j,x_i} h(s_j^{(2)})],
\end{align}
where $h$ is for encoding the semantic nodes and $\mathcal{N}_{x_i}^s=\{s_j\}_{j=1}^M$.  Finally, we append the numeric nodes to their corresponding semantic nodes as the representation of OCR tokens, denoted as $c = [c_1, ..., c_M]$. For OCR tokens which are not number-type, we concatenate a vector where the elements are all 0.

\subsection{Answer Prediction}
The answer prediction module takes the updated visual features $v = [v_1, ..., v_N]$ and OCR features $c = [c_1, ..., c_M]$ as inputs, and outputs the answer with copy mechanism~\cite{gu2016incorporating}. Concretely, first, the size of output space is extended to the vocabulary size + OCR number, where some indexes in the output space indicate copying the corresponding OCR as the answer, as shown in Fig.~\ref{model}. Then, we calculate the attention score on features of two modalities, and use attended features to generate the score of each answer, formulated as,
\begin{align}
y &= f_a([f_{\text{att}}^v(v, q)^T v; f_{\text{att}}^c(c, q)^T c]),
\end{align}
where $f_{\text{att}}^v$ and $f_{\text{att}}^c$ are Top-down attention networks in \cite{anderson2017bottom} and $f_a$ is an MLP to output the scores on all candidate answers. Finally, we optimize the binary cross entropy loss to train the whole network. This allows us to handle cases that the answer is in both the pre-defined answer space and the OCR tokens without penalizing for predicting either one. 

\section{Experiments}

\subsection{Experiments Setup}
\textbf{Datasets.}
We evaluate our model using the TextVQA dataset and Scene-Text VQA (ST-VQA) dataset.

For \textbf{TextVQA} dataset, it contains a total of 45,336 human-asked questions on 28,408 images from Open Image dataset~\cite{krasin2017openimages}. Each question-answer pair comes along with a list of tokens extracted by Object Character Recognition (OCR) models, Rosetta~\cite{borisyuk2018rosetta}. These questions are evaluated by VQA accuracy metric~\cite{goyal2016making}.

For \textbf{ST-VQA} dataset, it is composed of 23,038 images, paired with 31,791 human-annotated questions. In the Weakly Contextualized task of ST-VQA, a dictionary of 30,000 words are provided for all questions in this task; and the Open Dictionary task is open-lexicon. These questions are evaluated by two metrics, Average Normalized Levenshtein Similarity (ANLS)~\cite{levenshtein1966binary} and accuracy. 

\textbf{Implementation Details.}
For experiments on \textbf{TextVQA} dataset, we use answers that appear at least twice in the training set as our vocabulary. Thus, the size of our output space is the sum
of the vocabulary size and the OCR number, that is, $3997 + 50$. For question features, we use GloVe~\cite{pennington2014glove}, which is widely used in VQA models, to embed the words, then feed word embeddings to an LSTM~\cite{hochreiter1997long} with self-attention~\cite{yu2018beyond} to generate the question embedding. For encoding OCR tokens, GloVe only can represent out-of-vocabulary (OOV) words as 0-vectors which are not suitable for initialization them, so we use fastText~\cite{joulin2017bag}, which can represent OOV words as different vectors, to initialize OCR tokens. For image features, we use two kinds of pre-extracted visual features for each image provided by the TextVQA dataset, 1) 196 grid-based features obtained from pre-trained ResNet-152, and 2) 100 region-based features extracted from pre-trained Faster R-CNN model like~\cite{anderson2017bottom}. Both two visual features are 2048-dimensional. Note that, the Faster R-CNN provides the visual features of both objects and the scene texts because the detector produces an excessive amount of bounding boxes, where some bounding boxes will bound the scene texts.

Bounding box coordinates of objects and OCR tokens are first normalized into the interval of $[0, 1]$. Then we concatenate its center point, lower-left corner and upper-right corner’s coordinates, width, height, area, and aspect ratio, into a 10-dimensional feature. We used AdaMax optimizer~\cite{kingma2014adam} for optimization. A learning rate of 1e-2 is applied on all parameters except the fc7 layer for finetuning, which are trained with 5e-3.

For experiments on \textbf{ST-VQA} dataset, due to no available OCR results are provided, we use TextSpotter~\cite{he2018end} to extract scene text in images. For question and OCR token embedding, we use the same models as in TextVQA; and for image features, we only use Faster R-CNN features. Besides, we swap the prediction vocabulary to suit the change of the dataset. For Open Dictionary task, we collect answers which appear at least twice together with single-word answers which appear once in the training set as our vocabulary. For Weakly Contextualized task, given vocabulary of size 30,000 is directly utilized. Besides, the source codes are implemented with PyTorch~\cite{NEURIPS2019_9015}
\footnote{Our source codes are available at \url{http://vipl.ict.ac.cn/resources/codes}.}.

\begin{table}[]
\centering
\resizebox{6.3cm}{!}{%
\begin{tabular}{lcc}
\toprule
Method                & Val   & Test  \\ \midrule
Pythia                & 13.04 & 14.01 \\
LoRRA (BAN)           & 18.41 & -     \\
LoRRA (Pythia)        & 26.56 & 27.63 \\
BERT + MFH            & 28.96 & -     \\
\textbf{MM-GNN} (ours)                & \textbf{31.44} & \textbf{31.10} \\ \midrule
BERT + MFH (ensemble) & 31.50 & 31.44 \\
\textbf{MM-GNN} (ensemble) (ours)    & \textbf{32.92} & \textbf{32.46}      \\ \midrule
LA+OCR UB             & 67.56 & 68.24 \\ \bottomrule
\end{tabular}
}
\vspace{2pt}
\caption{VQA accuracy (\%) on the TextVQA dataset, comparison with baselines and state-of-the-art models. LA+OCR UB refers to maximum accuracy achievable by models using Large Vocabulary of LoRRA and OCR results provided by TextVQA dataset~\cite{singh2019towards}.}
\label{sota_textvqa}
\end{table}

\subsection{Results}
\textbf{Comparison with state-of-the-arts.}
Table~\ref{sota_textvqa} shows the comparison between our method and state-of-the-art approaches on validation and test set of TextVQA dataset. In the table, LoRRA (Pythia) is the baseline provided by TextVQA dataset~\cite{singh2019towards}. BERT + MFH is the winner of CVPR 2019 TextVQA challenge, which is considered as the state-of-the-art, and its results are quoted from its challenge winner talk. LA+OCR UB refers to maximum accuracy achievable by models using current OCR results and Large Vocabulary of LoRRA provided by TextVQA dataset~\cite{singh2019towards}. LoRRA and BERT+MFH utilize advanced fusion techniques to attend to OCR tokens which are encoded by pre-trained FastText~\cite{joulin2017bag}. BERT+MFH additionally introduces the powerful question encoder BERT~\cite{devlin2019bert} into the answering model. Our approach outperforms the above methods which mainly rely on pre-trained word embedding, and achieves state-of-the-art results. Table~\ref{sota_stvqa} compares our method and state-of-the-art approaches on Weakly Contextualized and Open Dictionary tasks of Scene-Text VQA datasets, where VTA is the winner model of ICDAR 2019 Competition on STVQA, which extends the Bottom-Up VQA model~\cite{anderson2017bottom} with BERT to encode the question and text. From the results, we can see that MM-GNN obtains an obvious improvement over baseline methods, e.g. SAN(CNN)+STR, and achieves comparable accuracies with VTA. 


\begin{table}[]
\centering
\resizebox{8.2cm}{!}{%
\begin{tabular}{lcccc}
\toprule
\multirow{2}{*}{Method}      & \multicolumn{2}{c}{Weakly Contextualized} & \multicolumn{2}{c}{Open Dictionary} \\ 
\cmidrule(lr){2-3} \cmidrule(lr){4-5}
                             & ANLS         & Acc.        & ANLS         & Acc.        \\ \midrule
SAAA                 & 0.085        & 6.36        & 0.085        & 6.36        \\
SAAA+STR             & 0.096        & 7.41        & 0.096        & 7.41        \\
SAN(LSTM)+STR        & 0.136        & 10.34       & 0.136        & 10.34       \\
SAN(CNN)+STR         & 0.135        & 10.46       & 0.135        & 10.46       \\  
VTA~\cite{Ali2019ICADAR} &  0.279   & 17.77       & 0.282        & 18.13   \\ \midrule
MM-GNN (ours)        & 0.203        & 15.69       & 0.207        & 16.00     \\ 
\bottomrule
\end{tabular}%
}
\vspace{2pt}
\caption{Average Normalized Levenshtein Similarity (ANLS) and accuracies (\%) of different methods on Weakly Contextualized and Open Dictionary tasks on the test set of ST-VQA dataset.}
\label{sota_stvqa}

\end{table}

\textbf{Effectiveness of Multi-Modal GNN.}
Our model's advantage lies in the introduction of a multi-modal graph and a well-designed message passing strategy between different sub-graphs to capture different types of contexts. Thus, we propose several variants of our model, where each variant ablates some aggregators to show their indispensability.
\begin{itemize}[itemsep= -2 pt,topsep = -1 pt]
    \item No-GNN: This variant directly uses the object and OCR token features extracted from pre-trained models to answer the questions without going through the multi-modal GNNs. Other modules (output, embeddings) are kept the same to MM-GNN.
    \item Vanilla GNN: This variant puts object and OCR token features in a single graph. It then performs an aggregation similar to Semantic-Semantic aggregator to update the representation of the nodes. Other modules are kept the same to MM-GNN.
    \item Combinations of VS, SS, and SN: These variants construct the multi-modal graph like MM-GNN, but only use one or two of the aggregators to update representations. VS, SS, and SN represent Visual-Semantic, Semantic-Semantic, and Semantic-Numeric aggregators respectively.
\end{itemize}


In addition, to better compare the results in detail, we categorize the questions in TextVQA into three types. The first type of question is \textbf{Unanswerable}, including questions that are unanswerable for given currently provided OCR tokens in TextVQA dataset. We obtain this type of question by checking whether the ground-truth answer absent from predefined answer vocabulary and provided OCR tokens. The second type of question has answers which can only be found in predefined answer vocabulary, such as ``red'', ``bus'',  and are not in OCR tokens, denoted as \textbf{Vocab}. The third type of question is \textbf{OCR} related questions where the answers derive from the OCR tokens. Due to \textbf{Unanswerable} type of questions cannot effectively evaluate the power of different variants, we report scores of \textbf{Vocab} and \textbf{OCR}, which are under the category of \textbf{Answerable}, and the \textbf{Overall} accuracy (including Unanswerable).

We evaluate the variants on the validation set of TextVQA dataset and report their accuracies on each type of question, as shown in Table~\ref{mm-gnn}. Comparing the performances of our full model MM-GNN with baseline No-GNN, we can see that MM-GNN outperforms NO-GNN with about 4\% on overall accuracy, and over 8\% on OCR related questions which are the main focus of TextVQA. 
This demonstrates that introducing the graph representation into TextVQA model can effectively help the answering procedure. Comparing the results of Vanilla GNN with MM-GNN series, we find that if message passing in GNN is not well-designed, directly applying GNN to TextVQA task is of little help. By comparing the results of SS, SN, and VS, we find that Visual-Semantic aggregator contributes most performance gain to OCR-related questions and overall accuracies. This demonstrates our idea that multi-modal contexts are effective in improving the quality of scene text representation.

However, we find that Numeric-Semantic aggregator contributes smaller than the other two aggregators, probably because the portion of questions querying the relations between the numbers, such as ``what is the largest number in the image?'', is relatively small. Thus, it limits the space to show the effectiveness of this aggregator.

\begin{table}[]
\centering
\resizebox{6.3cm}{!}{%
\begin{tabular}{lcccc}
\toprule
\multirow{2}{*}{Method} & \multicolumn{2}{c}{Answerable} & \multirow{2}{*}{Overall} \\ \cmidrule(lr){2-3}
            & Vocab          & OCR       &                          \\ \midrule
No-GNN    & 28.88           & 35.38      & 27.55           \\
Vanilla GNN & 28.29           & 37.70      & 28.58           \\ \midrule
VS & 27.54    & 41.38  & 30.14       \\ 
SS & \textbf{29.75}    & 38.89  & 29.71           \\
SN & 25.67    & 40.30  & 28.82      \\\midrule
VS + SS & 28.81    & 42.16  & 30.99      \\ 
VS + SN & 28.61        & 41.30      & 30.44        \\
SS + SN & 25.69        & 41.99      & 29.78        \\ \midrule
\textbf{VS + SS + SN} (ours)   & 27.85             & \textbf{43.36}      & \textbf{31.21}         \\ \bottomrule
\end{tabular}%
}
\vspace{2pt}
\caption{VQA accuracy (\%) of VQA models with different kinds of Graph Neural Networks on validation set of the TextVQA dataset.}
\label{mm-gnn}
\end{table}

\textbf{Impact of different combining methods.}
Choosing combination schemes controlling the fusion of a source node and the aggregated features of its neighboring nodes is one crucial part of Graph Neural Network design. Original \textbf{MM-GNN} is designed to gradually append additional information to each node to serve as hints to distinguish OCR tokens from each other and facilitate the answering model to locate the proper OCR token. Here, we replace our concatenation updater by several variants which are broadly used in other GNNs: 
\begin{itemize}[itemsep= -2 pt,topsep = -1 pt]
    \item Sum: this variant combines the features of source nodes and its neighboring nodes by sum operation, which is widely used in existing GNN works, such as~\cite{atwood2016diffusion}.
    \item Product: this variant updates each node by computing the element-wise multiplication of the node feature and aggregated features of its neighboring nodes.
    \item Concat + MLP: this variant updates each node by concatenating the node feature and aggregated features of its neighboring nodes, then uses an MLP to encode the concatenated features, which is used in previous visual language-related methods~\cite{hu2019language}.
\end{itemize}

\begin{table}[]
\centering
\resizebox{5.5cm}{!}{%
\begin{tabular}{lccc}
\toprule
\multirow{2}{*}{Method} & \multicolumn{2}{c}{Answerable} & \multirow{2}{*}{Overall} \\ \cmidrule(lr){2-3}
            & Vocab          & OCR       &                          \\ \midrule
Sum       & 27.40           & 40.40    & 29.59        \\
Product   & 27.89           & 32.18    & 25.79        \\ 
Concat+MLP & \textbf{28.11}           & 38.44    & 28.73        \\ \midrule
\textbf{Concat} (ours)              & 27.85      & \textbf{43.36}      & \textbf{31.21}      \\ \bottomrule
\end{tabular}
}
\vspace{2pt}
\caption{VQA accuracy (\%) of variants of MM-GNN with different combination schemes on validation set of TextVQA dataset.}
\label{aggregator}
\end{table}

\begin{figure}
\centering
\includegraphics[height=4.1cm]{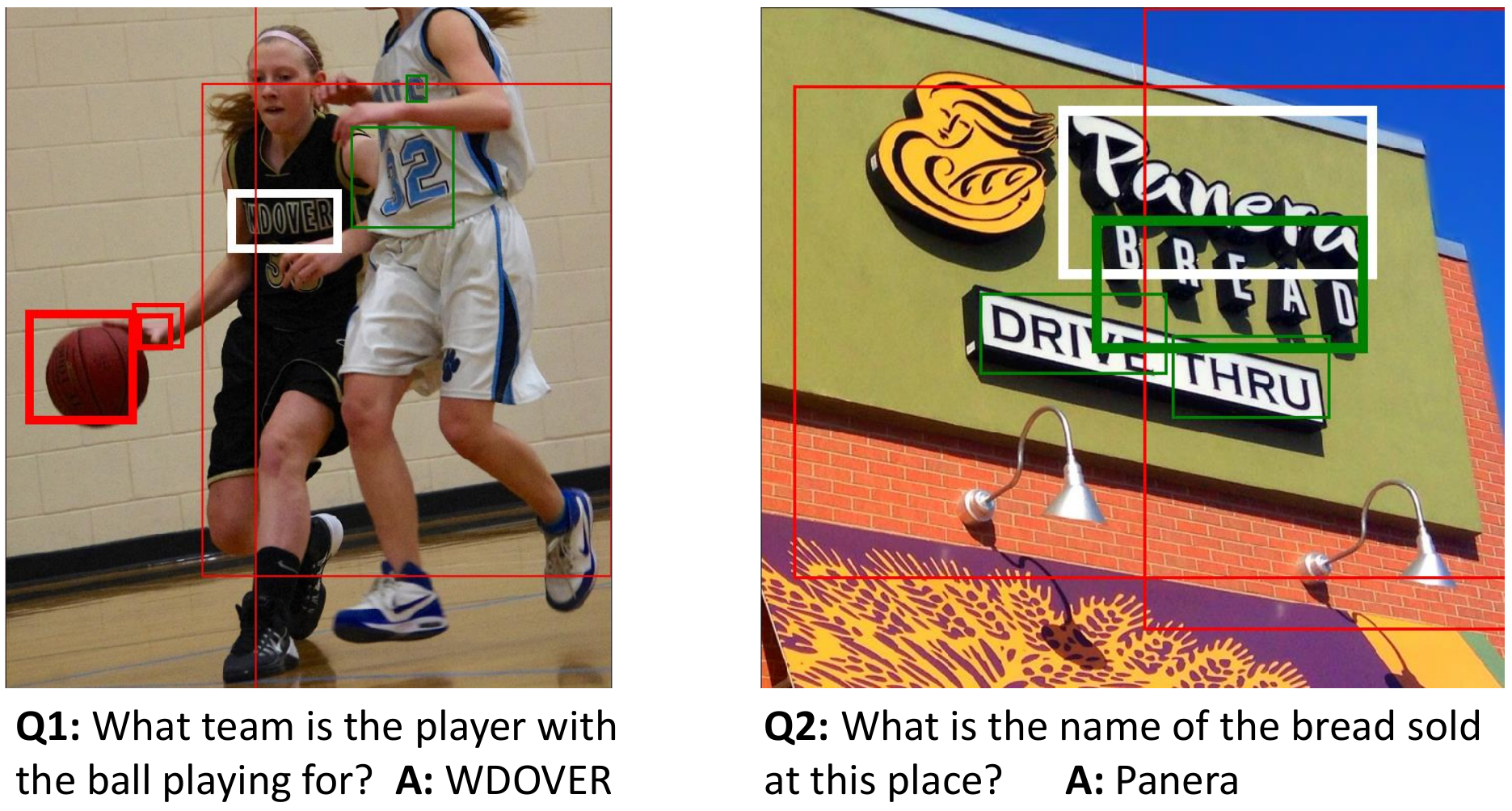}
\caption{
Visualization of attention results generated by \textbf{MM-GNN}. White boxes in the images bound the finally predicted OCR; red boxes are the objects most related to predicted OCR generated by Visual-Semantic aggregator; green boxes are the OCR tokens most related to predicted OCR generated by Semantic-Semantic aggregator. We only show boxes with attention value above a fixed threshold, with boxes more attended having thicker lines. It shows that, our attentions are sharp and truly attend on a few objects or texts that are relevant to answering questions.}
\label{attvisualization}
\end{figure}

\begin{figure*}
\centering
\includegraphics[height=11.2cm]{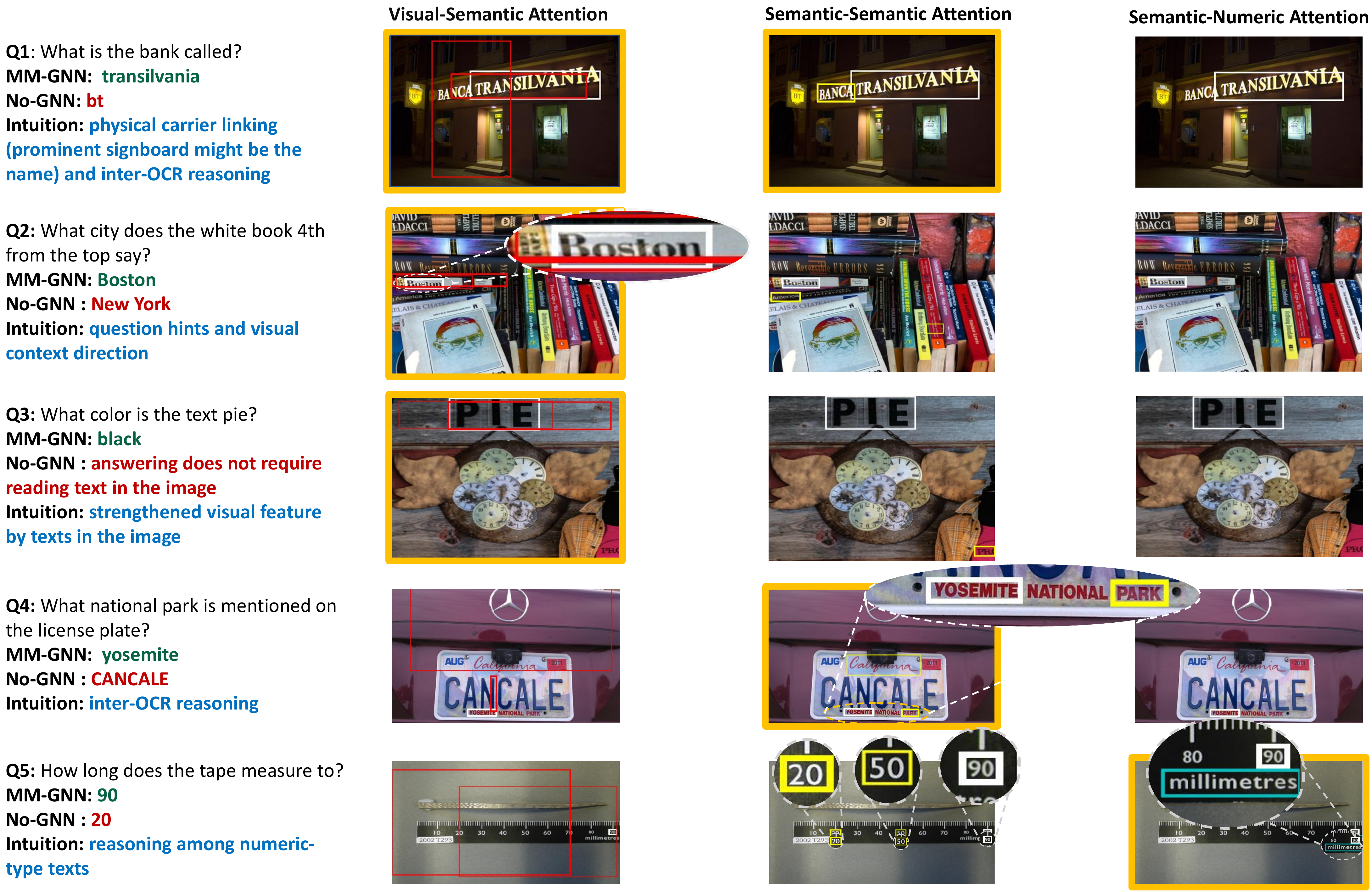}
\caption{Visualization of the reasoning procedure of MM-GNN model. We only display the attention from the OCR token selected as the answer in the answering module. The predicted OCR tokens are in \textbf{white} boxes. In the \textbf{Visual-Semantic Attention} column, we show the attention from OCR tokens to the most attended two visual objects, which are in \textbf{red} bounding boxes. The \textbf{Semantic-Semantic Attention} column displays attention between the predicted OCR token to the most attended OCR tokens, which are in \textbf{yellow} bounding boxes. In the \textbf{Semantic-Numeric Attention} column, the attentions from the predicted OCR token to other OCR tokens are shown (if any) in \textbf{cyan}. Images most important for answering the question are framed in \textbf{orange}, and the thickness of bounding boxes is proportional to their attention weights. These satisfying visualizations demonstrate that our model learns to do step-by-step reasoning in an explainable way.}
\label{mmgnn_vs_no_gnn}
\end{figure*}

We evaluate their performances on the validation set of TextVQA dataset, and the performances are shown in Table~\ref{aggregator}. We can see that all three schemes harm the performances more or less. Empirically, this is because the information between nodes and their neighborhoods are compressed, gradually averaging out the differences between node features, thereby bewildering the answering module when it tries to locate the question-related OCR token. Note that all three above combination schemes have the superiority of not changing node feature size through iterations; while our concatenation scheme looses this restriction to preserve more information in combination stage.

\subsection{Qualitative Analysis}
To gain an intuition of the attention distribution in aggregators, we visualized them in Fig.~\ref{attvisualization}. It shows that our model can produce very sharp attentions to do reasoning on graphs, and the attentions have good interpretability. In \textbf{Q1}, with the question querying about the player with a ball, OCR tokens are guided by attention module to incorporate more information related to the basketball; besides question hints, ``WDOVER'' naturally attends to the area of the player. In \textbf{Q2}, the OCR token ``Panera'' incorporates the position and semantic information from ``BREAD'' according to questions, and can be chosen in answering module because the model learns that the word above ``BREAD'' is very likely to be the name.

To better illustrate the answering procedure of the MM-GNN, we visualize the attention results of each aggregator when answering questions and compare the final answers of MM-GNN and baseline No-GNN. In Fig.~\ref{mmgnn_vs_no_gnn}, we show the results of several typical questions: \textbf{Q1} requires the model to utilize the visual context ``prominent texts on the signboard of a building'' to infer the semantic of unknown OCR token ``transilvania''. Besides, the OCR context ``banca'' also helps to find out that ``transilvania'' is the name of a bank. \textbf{Q2} requires to link the text ``Boston'' to its \emph{physical carriers} ``the white book'' and copy the OCR token as the answer,  \textbf{Q3} requires to link the text ``PIE'' to its \emph{physical carriers} ``the region containing black letters'', \textbf{Q4} requires to infer the semantic of the OCR token from its surrounding OCR tokens, and \textbf{Q5} evaluates the ability of finding the largest number. 

\section{Conclusion}
In this paper, we introduce a novel framework Multi-Modal Graph Neural Network (MM-GNN) for VQA with scene texts. The MM-GNN represents the image with multi-modal contents as a composition of three graphs, where each graph represents one modality. In addition, the designed multi-modal aggregators in MM-GNN utilize multi-modal contexts to obtain a finer representation of elements in the image, especially for unknown, rare or polysemous words. Experimentally, we show that our new image representation and message passing schemes greatly improve the performance of the VQA with scene texts and provide interpretable intermediate results.

\textbf{Acknowledgements.} This work is partially supported by Natural Science Foundation
of China under contracts Nos. 61922080, U19B2036, 61772500, and CAS Frontier Science Key Research Project No. QYZDJ-SSWJSC009.

{\small
\bibliographystyle{ieee_fullname}
\bibliography{egbib}

\begin{thebibliography}{10}\itemsep=-1pt

\bibitem{anderson2017bottom}
Peter Anderson, Xiaodong He, Chris Buehler, Damien Teney, Mark Johnson, Stephen
  Gould, and Lei Zhang.
\newblock Bottom-up and top-down attention for image captioning and vqa.
\newblock In {\em Proceedings of the IEEE Conference on Computer Vision and
  Pattern Recognition (CVPR)}, pages 6077--6086, 2018.

\bibitem{andreas2016learning}
Jacob Andreas, Marcus Rohrbach, Trevor Darrell, and Dan Klein.
\newblock Learning to compose neural networks for question answering.
\newblock In {\em Proceedings of NAACL-HLT}, pages 1545--1554, 2016.

\bibitem{andreas2016neural}
Jacob Andreas, Marcus Rohrbach, Trevor Darrell, and Dan Klein.
\newblock Neural module networks.
\newblock In {\em Proceedings of the IEEE Conference on Computer Vision and
  Pattern Recognition (CVPR)}, pages 39--48, 2016.

\bibitem{antol2015vqa}
Stanislaw Antol, Aishwarya Agrawal, Jiasen Lu, Margaret Mitchell, Dhruv Batra,
  C. Lawrence~Zitnick, and Devi Parikh.
\newblock Vqa: Visual question answering.
\newblock In {\em Proceedings of the IEEE International Conference on Computer
  Vision (ICCV)}, pages 2425--2433, 2015.

\bibitem{atwood2016diffusion}
James Atwood and Don Towsley.
\newblock Diffusion-convolutional neural networks.
\newblock In {\em Advances in Neural Information Processing Systems (NIPS)},
  pages 1993--2001, 2016.

\bibitem{ben2017mutan}
Hedi Ben-Younes, R{\'e}mi Cadene, Matthieu Cord, and Nicolas Thome.
\newblock Mutan: Multimodal tucker fusion for visual question answering.
\newblock In {\em Proceedings of the IEEE International Conference on Computer
  Vision (ICCV)}, pages 2612--2620, 2017.

\bibitem{Ali2019ICADAR}
Ali~Furkan Biten, Rub{\`{e}}n Tito, Andr{\'{e}}s Mafla, Llu{\'{\i}}s
  G{\'{o}}mez, Mar{\c{c}}al Rusi{\~{n}}ol, Minesh Mathew, C.~V. Jawahar, Ernest
  Valveny, and Dimosthenis Karatzas.
\newblock Icdar 2019 competition on scene text visual question answering.
\newblock {\em CoRR}, abs/1907.00490, 2019.

\bibitem{biten2019scene}
Ali~Furkan Biten, Ruben Tito, Andres Mafla, Lluis Gomez, Mar{\c{c}}al
  Rusi{\~n}ol, Ernest Valveny, C.~V. Jawahar, and Dimosthenis Karatzas.
\newblock Scene text visual question answering.
\newblock {\em Proceedings of the IEEE International Conference on Computer
  Vision (ICCV)}, pages 4291--4301, 2019.

\bibitem{borisyuk2018rosetta}
Fedor Borisyuk, Albert Gordo, and Viswanath Sivakumar.
\newblock Rosetta: Large scale system for text detection and recognition in
  images.
\newblock In {\em Proceedings of the 24th ACM SIGKDD International Conference
  on Knowledge Discovery \& Data Mining}, pages 71--79, 2018.

\bibitem{bruna2014spectral}
Joan Bruna, Wojciech Zaremba, Arthur Szlam, and Yann Lecun.
\newblock Spectral networks and locally connected networks on graphs.
\newblock In {\em International Conference on Learning Representations (ICLR)},
  2014.

\bibitem{defferrard2016convolutional}
Micha{\"e}l Defferrard, Xavier Bresson, and Pierre Vandergheynst.
\newblock Convolutional neural networks on graphs with fast localized spectral
  filtering.
\newblock In {\em Advances in Neural Information Processing Systems (NIPS)},
  pages 3844--3852, 2016.

\bibitem{devlin2019bert}
Jacob Devlin, Ming-Wei Chang, Kenton Lee, and Kristina Toutanova.
\newblock Bert: Pre-training of deep bidirectional transformers for language
  understanding.
\newblock In {\em Proceedings of the 2019 Conference of the North American
  Chapter of the Association for Computational Linguistics: Human Language
  Technologies, Volume 1 (Long and Short Papers)}, pages 4171--4186, 2019.

\bibitem{fout2017protein}
Alex Fout, Jonathon Byrd, Basir Shariat, and Asa Ben-Hur.
\newblock Protein interface prediction using graph convolutional networks.
\newblock In {\em Advances in Neural Information Processing Systems (NIPS)},
  pages 6530--6539, 2017.

\bibitem{fukui2016multimodal}
Akira Fukui, Dong~Huk Park, Daylen Yang, Anna Rohrbach, Trevor Darrell, and
  Marcus Rohrbach.
\newblock Multimodal compact bilinear pooling for visual question answering and
  visual grounding.
\newblock In {\em Conference on Empirical Methods in Natural Language
  Processing (EMNLP)}, pages 457--468, 2016.

\bibitem{girshick2015fast}
Ross Girshick.
\newblock Fast r-cnn.
\newblock In {\em Proceedings of the IEEE International Conference on Computer
  Vision (ICCV)}, pages 1440--1448, 2015.

\bibitem{goyal2016making}
Yash Goyal, Tejas Khot, Douglas Summers-Stay, Dhruv Batra, and Devi Parikh.
\newblock Making the v in vqa matter: Elevating the role of image understanding
  in visual question answering.
\newblock In {\em Proceedings of the IEEE Conference on Computer Vision and
  Pattern Recognition (CVPR)}, pages 6904--6913, 2017.

\bibitem{gu2016incorporating}
Jiatao Gu, Zhengdong Lu, Hang Li, and Victor~O.K. Li.
\newblock Incorporating copying mechanism in sequence-to-sequence learning.
\newblock In {\em In Annual Meeting of the Association for Computational
  Linguistics (ACL)}, pages 1631--1640, 2016.

\bibitem{he2018end}
Tong He, Zhi Tian, Weilin Huang, Chunhua Shen, Yu Qiao, and Changming Sun.
\newblock An end-to-end textspotter with explicit alignment and attention.
\newblock In {\em Proceedings of the IEEE Conference on Computer Vision and
  Pattern Recognition (CVPR)}, pages 5020--5029, 2018.

\bibitem{hochreiter1997long}
Sepp Hochreiter and J{\"u}rgen Schmidhuber.
\newblock Long short-term memory.
\newblock {\em Neural computation}, 9(8):1735--1780, 1997.

\bibitem{hu2017learning}
Ronghang Hu, Jacob Andreas, Marcus Rohrbach, Trevor Darrell, and Kate Saenko.
\newblock Learning to reason: End-to-end module networks for visual question
  answering.
\newblock In {\em Proceedings of the IEEE International Conference on Computer
  Vision (ICCV)}, pages 804--813, 2017.

\bibitem{hu2019language}
Ronghang Hu, Anna Rohrbach, Trevor Darrell, and Kate Saenko.
\newblock Language-conditioned graph networks for relational reasoning.
\newblock In {\em Proceedings of the IEEE International Conference on Computer
  Vision (ICCV)}, pages 10294--10303, 2019.

\bibitem{hudson2018compositional}
Drew~A. Hudson and Christopher~D. Manning.
\newblock Compositional attention networks for machine reasoning.
\newblock {\em International Conference on Learning Representations (ICLR)},
  2018.

\bibitem{hudson2019gqa}
Drew~A. Hudson and Christopher~D. Manning.
\newblock Gqa: A new dataset for real-world visual reasoning and compositional
  question answering.
\newblock In {\em Proceedings of the IEEE Conference on Computer Vision and
  Pattern Recognition (CVPR)}, pages 6700--6709, 2019.

\bibitem{johnson2017clevr}
Justin Johnson, Bharath Hariharan, Laurens van~der Maaten, Li Fei-Fei,
  C.~Lawrence Zitnick, and Ross Girshick.
\newblock Clevr: A diagnostic dataset for compositional language and elementary
  visual reasoning.
\newblock In {\em Proceedings of the IEEE Conference on Computer Vision and
  Pattern Recognition (CVPR)}, pages 1988--1997, 2017.

\bibitem{johnson2017inferring}
Justin Johnson, Bharath Hariharan, Laurens Van Der~Maaten, Judy Hoffman, Li
  Fei-Fei, C. Lawrence~Zitnick, and Ross Girshick.
\newblock Inferring and executing programs for visual reasoning.
\newblock In {\em Proceedings of the IEEE International Conference on Computer
  Vision (ICCV)}, pages 2989--2998, 2017.

\bibitem{joulin2017bag}
Armand Joulin, {\'E}douard Grave, Piotr Bojanowski, and Tom{\'a}{\v{s}}
  Mikolov.
\newblock Bag of tricks for efficient text classification.
\newblock In {\em Proceedings of the Conference of the European Chapter of the
  Association for Computational Linguistics (EACL)}, pages 427--431, 2017.

\bibitem{kazemzadeh2014referitgame}
Sahar Kazemzadeh, Vicente Ordonez, Mark Matten, and Tamara Berg.
\newblock Referitgame: Referring to objects in photographs of natural scenes.
\newblock In {\em Proceedings of the 2014 conference on empirical methods in
  natural language processing (EMNLP)}, pages 787--798, 2014.

\bibitem{kingma2014adam}
Diederik~P. Kingma and Jimmy Ba.
\newblock Adam: A method for stochastic optimization.
\newblock In {\em International Conference on Learning Representations (ICLR)},
  2015.

\bibitem{kipf2016semi}
Thomas~N. Kipf and Max Welling.
\newblock Semi-supervised classification with graph convolutional networks.
\newblock In {\em International Conference on Learning Representations (ICLR)},
  2017.

\bibitem{krasin2017openimages}
Ivan Krasin, Tom Duerig, Neil Alldrin, Vittorio Ferrari, Sami Abu-El-Haija,
  Alina Kuznetsova, Hassan Rom, Jasper Uijlings, Stefan Popov, Andreas Veit,
  Serge Belongie, Victor Gomes, Abhinav Gupta, Chen Sun, Gal Chechik, David
  Cai, Zheyun Feng, Dhyanesh Narayanan, and Kevin Murphy.
\newblock Openimages: A public dataset for large-scale multi-label and
  multi-class image classification.
\newblock {\em Dataset available from https://github.com/openimages}, 2017.

\bibitem{levenshtein1966binary}
Vladimir~I. Levenshtein.
\newblock Binary codes capable of correcting deletions, insertions, and
  reversals.
\newblock In {\em Soviet Physics Doklady}, volume~10, pages 707--710, 1966.

\bibitem{li2019relation}
Linjie Li, Zhe Gan, Yu Cheng, and Jingjing Liu.
\newblock Relation-aware graph attention network for visual question answering.
\newblock In {\em Proceedings of the IEEE International Conference on Computer
  Vision (ICCV)}, pages 10313--10322, 2019.

\bibitem{lu2016hierarchical}
Jiasen Lu, Jianwei Yang, Dhruv Batra, and Devi Parikh.
\newblock Hierarchical question-image co-attention for visual question
  answering.
\newblock In {\em Advances in Neural Information Processing Systems (NIPS)},
  pages 289--297, 2016.

\bibitem{malinowski2014multi}
Mateusz Malinowski and Mario Fritz.
\newblock A multi-world approach to question answering about real-world scenes
  based on uncertain input.
\newblock In {\em Advances in Neural Information Processing Systems (NIPS)},
  pages 1682--1690, 2014.

\bibitem{noh2016image}
Hyeonwoo Noh, Paul Hongsuck~Seo, and Bohyung Han.
\newblock Image question answering using convolutional neural network with
  dynamic parameter prediction.
\newblock In {\em Proceedings of the IEEE Conference on Computer Vision and
  Pattern Recognition (CVPR)}, pages 30--38, 2016.

\bibitem{norcliffe2018learning}
Will Norcliffe-Brown, Stathis Vafeias, and Sarah Parisot.
\newblock Learning conditioned graph structures for interpretable visual
  question answering.
\newblock In {\em Advances in Neural Information Processing Systems (NIPS)},
  pages 8334--8343, 2018.

\bibitem{NEURIPS2019_9015}
Adam Paszke, Sam Gross, Francisco Massa, Adam Lerer, James Bradbury, Gregory
  Chanan, Trevor Killeen, Zeming Lin, Natalia Gimelshein, Luca Antiga, Alban
  Desmaison, Andreas Kopf, Edward Yang, Zachary DeVito, Martin Raison, Alykhan
  Tejani, Sasank Chilamkurthy, Benoit Steiner, Lu Fang, Junjie Bai, and Soumith
  Chintala.
\newblock Pytorch: An imperative style, high-performance deep learning library.
\newblock pages 8024--8035, 2019.

\bibitem{pennington2014glove}
Jeffrey Pennington, Richard Socher, and Christopher Manning.
\newblock Glove: Global vectors for word representation.
\newblock In {\em Proceedings of the 2014 conference on empirical methods in
  natural language processing (EMNLP)}, pages 1532--1543, 2014.

\bibitem{ren2015exploring}
Mengye Ren, Ryan Kiros, and Richard Zemel.
\newblock Exploring models and data for image question answering.
\newblock In {\em Advances in Neural Information Processing Systems (NIPS)},
  pages 2953--2961, 2015.

\bibitem{scarselli2008graph}
Franco Scarselli, Marco Gori, Ah~Chung Tsoi, Markus Hagenbuchner, and Gabriele
  Monfardini.
\newblock The graph neural network model.
\newblock {\em IEEE Transactions on Neural Networks}, 20(1):61--80, 2008.

\bibitem{schwartz2017high}
Idan Schwartz, Alexander Schwing, and Tamir Hazan.
\newblock High-order attention models for visual question answering.
\newblock In {\em Advances in Neural Information Processing Systems (NIPS)},
  pages 3667--3677, 2017.

\bibitem{shah2019kvqa}
Sanket Shah, Anand Mishra, Naganand Yadati, and Partha~Pratim Talukdar.
\newblock Kvqa: Knowledge-aware visual question answering.
\newblock In {\em Proceedings of the AAAI Conference on Artificial Intelligence
  (AAAI)}, pages 8876--8884, 2019.

\bibitem{shi2019explainable}
Jiaxin Shi, Hanwang Zhang, and Juanzi Li.
\newblock Explainable and explicit visual reasoning over scene graphs.
\newblock In {\em Proceedings of the IEEE Conference on Computer Vision and
  Pattern Recognition (CVPR)}, pages 8376--8384, 2019.

\bibitem{singh2019towards}
Amanpreet Singh, Vivek Natarajan, Meet Shah, Yu Jiang, Xinlei Chen, Dhruv
  Batra, Devi Parikh, and Marcus Rohrbach.
\newblock Towards vqa models that can read.
\newblock In {\em Proceedings of the IEEE Conference on Computer Vision and
  Pattern Recognition (CVPR)}, pages 8317--8326, 2019.

\bibitem{teney2017graph}
Damien Teney, Lingqiao Liu, and Anton van~den Hengel.
\newblock Graph-structured representations for visual question answering.
\newblock In {\em Proceedings of the IEEE International Conference on Computer
  Vision (ICCV)}, pages 804--813, 2017.

\bibitem{velivckovic2017graph}
Petar Veli{\v{c}}kovi{\'c}, Guillem Cucurull, Arantxa Casanova, Adriana Romero,
  Pietro Lio, and Yoshua Bengio.
\newblock Graph attention networks.
\newblock In {\em International Conference on Learning Representations (ICLR)},
  2017.

\bibitem{wang2019neighbourhood}
Peng Wang, Qi Wu, Jiewei Cao, Chunhua Shen, Lianli Gao, and Anton van~den
  Hengel.
\newblock Neighbourhood watch: Referring expression comprehension via
  language-guided graph attention networks.
\newblock In {\em Proceedings of the IEEE Conference on Computer Vision and
  Pattern Recognition (CVPR)}, pages 1960--1968, 2019.

\bibitem{wang2018fvqa}
Peng Wang, Qi Wu, Chunhua Shen, Anthony Dick, and Anton van~den Hengel.
\newblock Fvqa: Fact-based visual question answering.
\newblock {\em IEEE Transactions on Pattern Analysis and Machine Intelligence},
  40(10):2413--2427, 2018.

\bibitem{xu2018powerful}
Keyulu Xu, Weihua Hu, Jure Leskovec, and Stefanie Jegelka.
\newblock How powerful are graph neural networks?
\newblock In {\em International Conference on Learning Representations (ICLR)},
  2019.

\bibitem{yang2016stacked}
Zichao Yang, Xiaodong He, Jianfeng Gao, Li Deng, and Alex Smola.
\newblock Stacked attention networks for image question answering.
\newblock In {\em Proceedings of the IEEE Conference on Computer Vision and
  Pattern Recognition (CVPR)}, pages 21--29, 2016.

\bibitem{yi2018neural}
Kexin Yi, Jiajun Wu, Chuang Gan, Antonio Torralba, Pushmeet Kohli, and Josh
  Tenenbaum.
\newblock Neural-symbolic vqa: Disentangling reasoning from vision and language
  understanding.
\newblock In {\em Advances in Neural Information Processing Systems (NIPS)},
  pages 1039--1050, 2018.

\bibitem{yu2018beyond}
Zhou Yu, Jun Yu, Chenchao Xiang, Jianping Fan, and Dacheng Tao.
\newblock Beyond bilinear: Generalized multimodal factorized high-order pooling
  for visual question answering.
\newblock {\em IEEE Transactions on Neural Networks and Learning Systems},
  29(12):5947--5959, 2018.

\bibitem{zellers2019recognition}
Rowan Zellers, Yonatan Bisk, Ali Farhadi, and Yejin Choi.
\newblock From recognition to cognition: Visual commonsense reasoning.
\newblock In {\em Proceedings of the IEEE Conference on Computer Vision and
  Pattern Recognition (CVPR)}, pages 6720--6731, 2019.

\bibitem{zhang2016yin}
Peng Zhang, Yash Goyal, Douglas Summers-Stay, Dhruv Batra, and Devi Parikh.
\newblock Yin and yang: Balancing and answering binary visual questions.
\newblock In {\em Proceedings of the IEEE Conference on Computer Vision and
  Pattern Recognition (CVPR)}, pages 5014--5022, 2016.

\end{thebibliography}
}

\clearpage



\begin{onecolumn}
\centerline{\large \textbf{Supplementary Material}}
    \section*{Overview}
    In the supplementary material, we display more implementation details of number encoder in graph construction (Sec.3.1 of the main body) and more experimental results about our method Multi-Modal Graph Neural Network (MM-GNN), including quantitative (on Sec.~\ref{secA} ) and qualitative analysis (on Sec.~\ref{secB}).
    
    \section{Number encoder}
    In MM-GNN, digital numbers and days of the week which are similar to time-related strings containing periodic information, e.g. “Sunday”, are considered as numeric type strings. Besides, for periodic numbers, e.g. 10:00, we first normalize it into $\frac{10}{24}$, then we apply the cosine embedding function $x \rightarrow cos\, 2\pi x$. Note that, how to encode the numeric strings is still an open problem, different encoders can capture different relations between numbers, e.g. an alternative representation that uses the polar coordinate system which uses two functions cosine and sine to encode a number could be better in representing the periodic numbers.

    \section{Quantitative Analysis}
    \label{secA}
    
    \textbf{The impact of the order of aggregators.}
    In Multi-Modal Graph Neural Network (MM-GNN), the three aggregators which update the representations of nodes in different sub-graphs are performed in a particular order, that is, first perform Visual-Semantic aggregator (VS), then perform Semantic-Semantic aggregator (SS), finally Semantic-Numeric aggregator (SN). In this part, we evaluate the influences of all different orders of aggregators. The results are shown in Table~\ref{order}. 

    From the results, we can see that the performances of different variants are similar to each other, which indicates that our proposed MM-GNN is robust to changes in the order. This probably thanks to the functions of three aggregators have relatively low dependencies on each other. 
    \begin{table}[hbtp]
    \centering
    \resizebox{5.9cm}{!}{%
    \begin{tabular}{lccc}
    \toprule
    \multirow{2}{*}{Method} & \multicolumn{2}{c}{Answerable} & \multirow{2}{*}{Overall} \\ \cmidrule(lr){2-3}
                & Vocab          & OCR       &                          \\ \midrule
    SS-VS-SN  & 26.71 &  42.99    & 30.54        \\
    SS-SN-VS  & 26.88 &  43.11    & 30.65        \\
    VS-SN-SS  & 25.80 &  43.08    & 30.27        \\
    SN-SS-VS  & 26.58 &  42.97    & 30.46        \\
    SN-VS-SS  & 27.66 &  41.63    & 30.33        \\ \midrule
    VS-SS-SN (ours)    & \textbf{27.85} & \textbf{43.36}      & \textbf{31.21}      \\ \bottomrule
    \end{tabular}
    }
    \vspace{2pt}
    \caption{VQA accuracy (\%) of variants of MM-GNN with different aggregators order on validation set of TextVQA dataset.}
    \label{order}
    \end{table}

    \textbf{Results on different question types.}
    Similar to VQA dataset~\cite{antol2015vqa}, we categorize the questions in TextVQA into three groups based on their question-type, i.e., yes/no, number and others. The performances of MM-GNN and baseline No-GNN on different question types are shown in Table~\ref{type}. We can see that our method mainly outperforms the baseline on others-type questions as expected, because these questions are mostly related to understanding diverse scene texts.

    \begin{table}[bhpt]
        \centering
        \vspace{-0.2cm}
        \resizebox{6.3cm}{!}{
        \begin{tabular}{lcccc}
        \toprule
        Model  & yes/no & number & others & Final \\ \midrule
        No-GNN & 88.79  & 35.14  & 22.65  & 27.55 \\
        MM-GNN & \textbf{88.93}  & \textbf{36.13}  & \textbf{27.36}  & \textbf{31.21} \\ \bottomrule
        \end{tabular}%
        }
        \vspace{2pt}
        \caption{VQA accuracy (\%) of MM-GNN with on different types questions of TextVQA dataset compared to No-GNN.}
        \label{type}
    \end{table}

    \section{Qualitative Analysis}
    \label{secB}
    In Fig.~\ref{success}, we show more successful cases for our MM-GNN model on TextVQA dataset. We show that MM-GNN obtains the correct answer, along with reasonable attention results when utilizing the multi-modal contexts. In Fig.~\ref{failure}, we show some failure cases of MM-GNN and analyze the possible reason for each example in the image.

    \clearpage
    \clearpage
  
    \begin{figure*}[t]
    \centering
    \includegraphics[height=20.5cm]{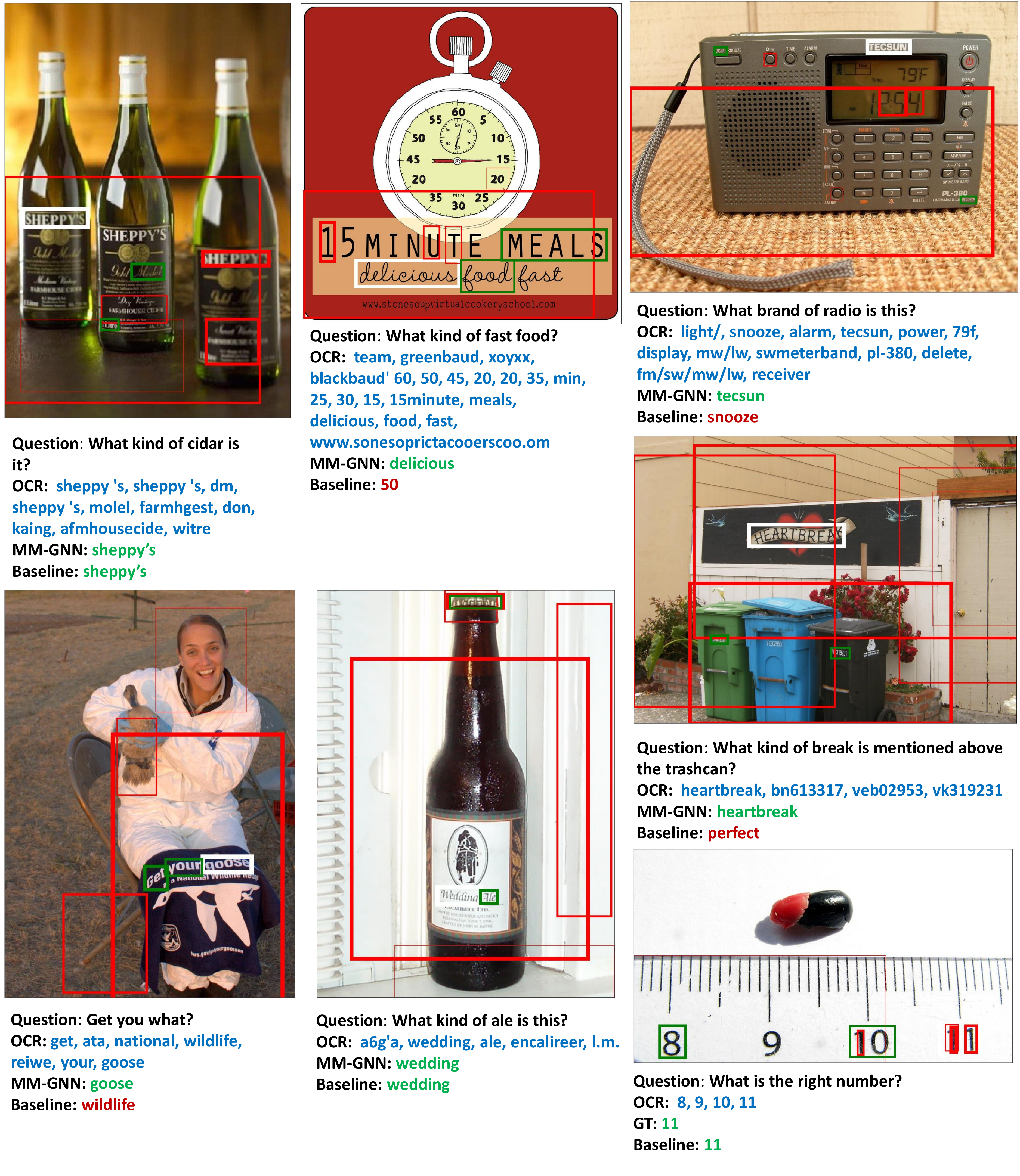}
    \caption{
    \textbf{Successful case analysis}. The predicted OCR is bounded in a white box. We show the attention from the predicted OCR  tokens  to  the  most  attended  five  visual  objects in Visual-Semantic aggregator (in red bounding boxes) and the attention between OCR tokens to the most attended two OCR tokens in Semantic-Semantic aggregator (in green bounding boxes), where bolder bounding box indicates higher attention value.}
    \label{success}
    \vspace{-0.5cm}
    \end{figure*}

    \clearpage

    \begin{figure*}[ht]
    \centering
    \includegraphics[height=20.5cm]{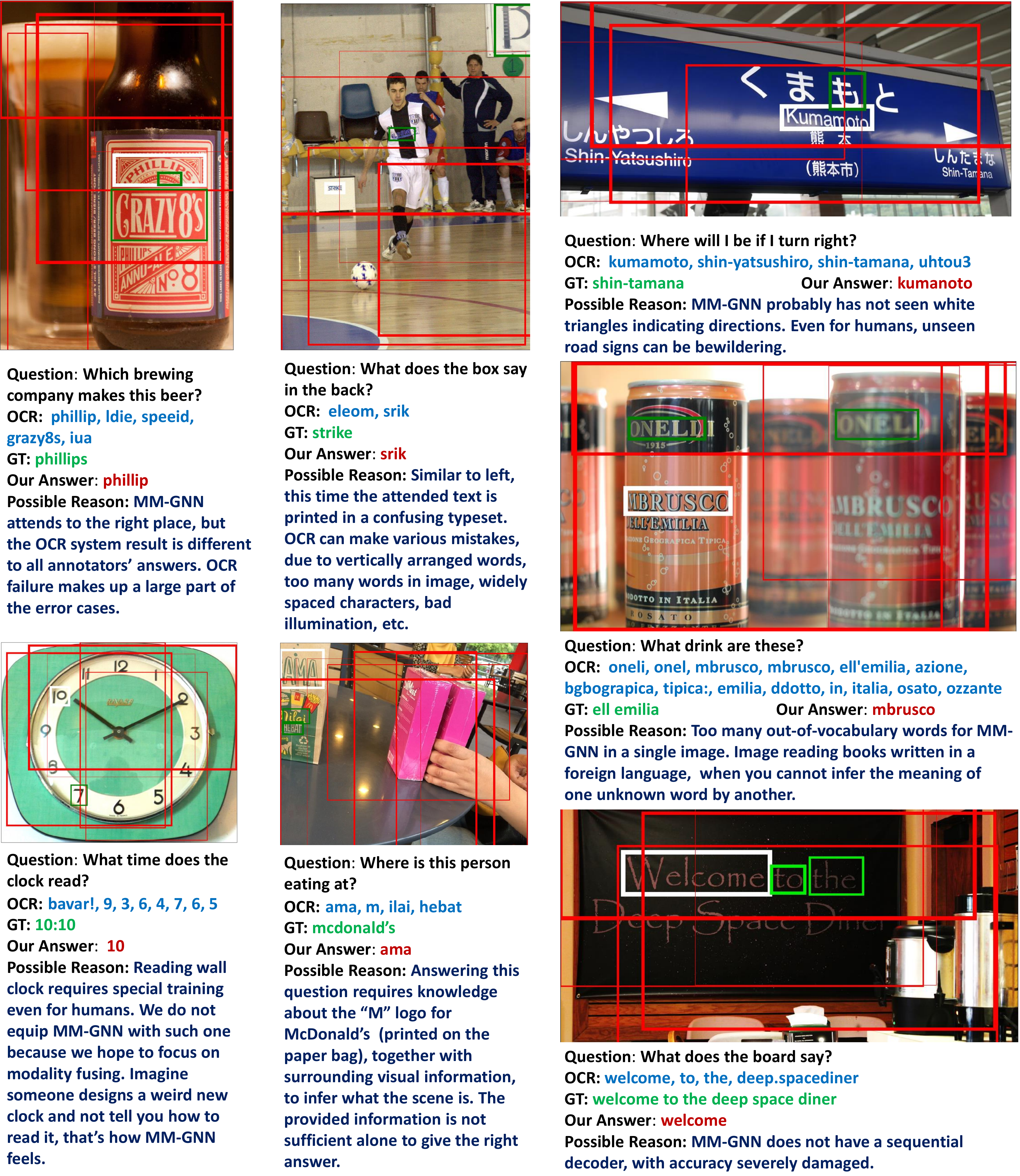}
    \caption{
    \textbf{Failure case analysis}. The predicted OCR is bounded in a white box. We show the attention from  OCR  tokens to the most attended five visual objects in Visual-Semantic aggregator (in red bounding boxes) and the attention between OCR tokens to the most attended two OCR tokens in Semantic-Semantic aggregator (in green bounding boxes), where bolder bounding box indicates higher attention value.}
    \label{failure}
    \vspace{-0.5cm}
    \end{figure*}
\end{onecolumn}
\clearpage

\end{document}